\pdfoutput=1
\documentclass{article}

\usepackage[final]{nips_2017}

\usepackage[dvipsnames,svgnames,x11names,hyperref]{xcolor}

\newcommand{\xxnote}[3]{}
\ifx\hidenotes\undefined
  \renewcommand{\xxnote}[3]{\color{#2}{#1: #3}}
\fi

\usepackage[utf8]{inputenc} 
\usepackage[T1]{fontenc}    
\usepackage{hyperref}       
\usepackage{url}            
\usepackage{booktabs}       
\usepackage{amsfonts}       
\usepackage{nicefrac}       
\usepackage{microtype}      
\usepackage{epigraph}
\usepackage{graphicx}
\graphicspath{ {images/}}
\usepackage{array}

\title{Incomplete Contracting and AI Alignment}
  
\author{
   Dylan Hadfield-Menell \\
 Department of Electrical Engineering and Computer Science\\
   University of California \\
   Berkeley, CA 94720\\
OpenAI, Center for Human-Compatible Artificial Intelligence\\
   \texttt{dhm@cs.berkeley.edu}
  \And Gillian K. Hadfield \\
Law School and Department of Economics\\
  University of Southern California\\
  Los Angeles, CA 90089\\ 
  Center for Human-Compatible Artificial Intelligence\\
  \texttt{ghadfield@law.usc.edu} 
   }
   
\begin{document}

\maketitle

\epigraph{There can never be complete communication between two people; a promise made and a promise heard are two different things...Thus [promises] can never be a \emph{complete} basis for dealing with the future.}{Ian R. Macneil, \emph{The Many Futures of Contracts, 1974} }

\section{Introduction}
When we design and deploy an AI agent, we specify what we want it to do. In reinforcement learning, for example, we specify a reward function, which tells the agent the value of all state and action combinations. Good algorithms then generate AI behavior that performs well according to this reward function. The AI alignment problem arises because of differences between the specified reward function and what relevant humans (the designer, the user, others affected by the agent's behavior) actually value.  AI researchers intend for their reward functions to give the correct rewards in all states of the world so as to achieve the objectives of relevant humans.  But often AI reward functions are---unintentionally and unavoidably---misspecified. They may accurately reflect human rewards in the circumstances that the designer thought about but fail to accurately specify how humans value all state and action combinations. When OpenAI decided to reward a reinforcement learning system with points earned in a boat-racing video game, for example, they didn't anticipate that the system would figure out that it could score high points by spinning in circles around points-earning targets in the course and not finishing the race at all \citep{OpenAI:2016}. \


Although the AI alignment problem is relatively new for computer scientists, it has a clear analogue in the human principal-agent problem long studied by economists and legal scholars. In these settings a human agent is tasked with taking actions that achieve a principal's objectives. The ideal way to align principal and agent is to design a \textit{complete contingent contract} \citep{Williamson:1975}. This is an enforceable agreement that specifies the reward received by the agent for all actions and states of the world. The contract could be enforced by monetary transfers or punishments imposed by a coercive institution, such as a court. Or it could be enforced by a private actor or group of actors who penalize contract violations by, for example, imposing social sanctions or cutting off valuable relationships; the latter contracts are known as \textit{relational contracts} \citep{Macaulay:1963,Macneil:1974,Levin:2003,Gil:2017}. A complete contingent contract implements desired behavior by the agent in all states of the world. It perfectly aligns the agent's incentives with the principal's objective. 

For several decades economists and legal scholars have recognized that writing complete contracts is routinely impossible for a wide variety of reasons \citep{Macneil:1974,Williamson:1975,Shavell:1984,Tirole:1999,aghion2011incomplete}. States of the world (particularly with respect to private information held by principal or agent) may be \textit{non-contractible}: unobservable or, if observable, unverifiable by contract enforcers. Humans are boundedly rational \citep{Williamson:1975} and may not be able to evaluate the optimal actions in all states of the world or determine optimal incentives. They may not be able to predict or even imagine all possible states of the world. It may not be possible unambiguously to describe states of the world and optimal actions. Even if none of these limits are hit, it may simply be too costly to write this all down in a manner that can be enforced at reasonable cost. And even if a contract looks complete--"deliver 100 pound of peas next Tuesday and I'll pay you \$200"--it may fail in fact to capture the parties' intent--not to require delivery if there is a hurricane that wipes out the pea crop or payment if the peas are rotten, for example. Because the contract does not specify the intended behavior in all contingencies it is incomplete. Contracts in human relationships are usually, and maybe necessarily, incomplete.  \

In this paper, we suggest that the analysis of incomplete contracting developed by law and economics researchers can provide a useful framework for understanding the AI alignment problem and help to generate a systematic approach to finding solutions. We first provide an overview of the incomplete contracting literature and explore parallels between this work and the problem of AI alignment. As we emphasize, misalignment between principal and agent is a core focus of economic analysis. We highlight some technical results from the economics literature on incomplete contracts that may provide insights for AI alignment researchers.  Our core contribution, however, is to bring to bear an insight that economists have been urged to absorb from legal scholars and other behavioral scientists: the fact that human  contracting is supported by substantial amounts of external structure, such as generally available institutions (culture, law) that can supply implied terms to fill the gaps in incomplete contracts. We propose a research agenda for AI alignment work that focuses on the problem of how to build AI that can replicate the human cognitive processes that connect individual incomplete contracts with this supporting external structure.   
\section{The Fundamental Problem of Misalignment}
Economies are built on specialization and the division of labor--meaning that the production and allocation of different things of value needs to be coordinated across a group of humans. The challenge of aligning the interests of one actor with others' is at the core of modern economic theory.  The first of two fundamental welfare theorems \citep{Arrow:1951,Debreu:1959} states that if markets are complete (all possible trades can be made including those that involve future goods and services and third-party effects) and perfectly competitive (there are no transaction costs, all information is common knowledge, and no one holds market power), then voluntary trading in a market economy will produce a result that is Pareto efficient:  there is no way to reallocate resources or goods so as to make someone better off without making someone else worse off.  This is a form of alignment:  the market outcome aligns with the solution that maximizes a social welfare function that aggregates the values of all members of the economy, weighted by their initial endowments of goods.  (The weights come from the choice to measure social welfare by the Pareto criterion: someone who has little to begin with has a low weight in the social welfare function maximized by market trade.) The second welfare theorem then states that for any distributive goal--for any final distribution of goods or social welfare weights that society chooses--there is an initial allocation of endowments (including labor) such that a perfectly competitive and complete market economy will produce that final distribution.  In theory, perfectly competitive and complete markets serve as a mechanism to align individual decisions about production and trade so as to maximize a social welfare function.  

Despite the centrality of these core welfare theorems, most work in economics focuses on the failure of markets to be perfectly competitive and complete. The welfare theorems for perfect markets merely provide a framework for thinking about how to design production and allocation systems--markets, organizations, laws--to achieve better outcomes from a social welfare point of view. Departures from perfect and complete markets introduce costs due to distortion, that is, a failure of alignment. Some things that humans value, such as rights to fair treatment or caring for one's own children \citep{Hadfield:2005}, cannot be fully traded . Most fundamentally, there is no coherent social welfare function that is based exclusively on subjective assessments of own utility; any coherent social welfare function requires collective judgments to be made about what values to pursue \citep{Arrow:1951,Sen:1985} and so there is inevitable ``misalignment'' with the values of some humans.

Misalignment in the human principal agent setting is responsible for the economic loss associated with delegation of decisions over productive effort. In a perfect frictionless world, where all factors that affect outcomes are common knowledge and any agreement between actors can be costlessly written and enforced, voluntary agreements in which an agent agrees to take certain actions and a principal agrees to compensate the agent in particular ways will align the interests of agent and principal. Core results in the theory of contracts then explore whether it is possible to align interests (achieve the "first-best" promised by the fundamental welfare theorems) when there is hidden information such as when an agent has private information about the cost of taking an action (adverse selection) or about the action chosen (moral hazard) \citep{Laffont:1989}. Sometimes this is indeed possible. If an agent is privately informed about the cost of a project, and is risk-neutral and not wealth-constrained, the first-best is achievable by effectively selling the project to the agent. The agent pays the principal a lump-sum amount (for example, equal to the full surplus generated by efficient production choices by the agent) and then has the right to collect all the realized returns to the project; the agent then faces an incentive to choose the most productive action. But in general, the first-best is not achievable: the optimal contract trades off giving the agent incentives to be productive (the size of the pie) and the achievement of the principal's goal or utility (share of the pie.) 

Misalignment in the design of artificially intelligent agents can be thought of in parallel terms. AI designers, like contract designers, are faced with the challenge of achieving intended goals in light of the limitations that arise from translating those goals into implementable structures to guide agent behavior (learning algorithms and reward functions). An AI is misaligned whenever it chooses behaviors based on a reward function that is different from the true welfare of relevant humans. We see misalignment as the general description of a wide variety of problems that go by different names in AI research. \cite{Amodei:2016} collect a set of cases that they refer to as ``accidents'': situations in which a human designer has an objective in mind but the system as designed and deployed produces ``harmful and unexpected results.'' They propose several mechanisms producing such accidents: negative side-effects, reward hacking, limited capacity for human oversight, differences between training and deployment environments, and uncontrolled or unexpected exploration after deployment. Alignment problems also arise because of the difficulty of representing and implementing human values. The problems of fairness and bias in machine learning algorithms are fundamentally alignment problems. The technical literature here (see, e.g. \citep{Zemel:2013,Lum:2016,Hardt:2016,Kleinberg:2016,Zafar:2016}) seeks to develop techniques to align algorithmic decisions with complex human goals such as discriminating between prospective employees on ability but not gender or race. AI safety problems such as safe interruptibility \citep{Orseau:2016}, the off-switch game \citep{Hadfield-Menell:2017a} and corrigibility \citep{soares2015corrigibility} are also alignment problems: these are efforts to ensure that AI agents value shut down or modification of their reward functions in the same way that humans do or at least are indifferent to such efforts. And at the most general level, the question of how to elicit and aggregate preferences when there are multiple humans affected by the behavior of an artificial agent \citep{Rossi:2011} is an alignment problem.  Indeed, it is the basic alignment problem addressed by the fundamental theorems of welfare economics.  


\section{Reasons for Misalignment}
It is natural to think that misalignment between agent and principal is just an error in design. And indeed, sometimes misalignment in the human principal agent setting is the result of bad contract drafting and sometimes in the context of artificial intelligence it is the result of straightforward misspecification of what the designer wants.  But these are not the particularly interesting or challenging cases of misalignment. Indeed, this is what is fundamental in the economic framework: complete contracting is routinely not possible, so we analyze optimal incomplete contracts.  Similarly, we suggest that perfect reward specification is routinely not possible, so AI researchers should be focused on designing optimally in the face of irredeemable divergence between AI and human utility. In this section we briefly collect the reasons for contract incompleteness and then provide what we think are the parallel reasons for reward misspecification in AI. 

The most commonly cited reason that contracts are incomplete is because completeness is practically impossible or costly:
\begin{itemize}
    \item \emph{unintended incompleteness}: contract designers fail to identify all circumstances that affect the value of the contract. This is sometimes referred to as bounded rationality \citep{simon1955behavioral,Williamson:1975}. 
    \item \emph{economizing on costly cognition and drafting}: contract designers choose not to invest in the costly cognitive effort of discovering, evaluating and drafting contract terms to cover all circumstances \citep{Shavell:1980,Shavell:1984,ScottTriantis:2006}
     \item \emph{economizing on enforcement costs}: contract designers leave out terms that are costly to enforce because they require more or more costly evidence or because the potential for disputes and/or court errors increases with the complexity of the contract \citep{Klein:1980, Schwartz:2003, Halonen:2013}
    \item \emph{non-contractibility}: some contingencies and/or actions are left out because either they cannot be observed or, even if observable, they cannot be verified by enforcers at reasonable cost. This might be because of hidden information or it might be because it is not possible to communicate (describe) the contingency or action in unambiguous terms \citep{Grossman:1986, Maskin:1999a}
\end{itemize}    
These reasons for incompleteness seem to  us reasonably to translate over to the AI context.  Rewards may fail to address all relevant circumstances because designers simply did not (and perhaps could not) think of everything. (See, e.g., \cite{Hadfield-Menell:2017b}. Designers may have deliberately chosen not to invest additional effort to identify or code for possible contingencies. And some reward structures are, given the state of the art, simply not implementable.  If we think of a contract as an implemented reward structure for a human agent, the analog to non-contractibility is a learning problem that is not solvable with known techniques.  Costly enforcement in the human contracting context ultimately leads to non-contractibility; the AI analog to incompleteness due to costly enforcement is an unaligned reward function as a result of engineering limitations on what rewards we can successfully deploy for a learning agent.  Limitations on feasible specified rewards has led, for example, to techniques of inverse reinforcement learning \citep{Ng:2000} and approaches in which a robot predicts the reward function from human feedback on a subset of actions \citep{Christiano:2017, dorsa2017active, daniel2014active}.\

In the above accounts, contractual incompleteness is undesirable: if humans had complete shared information, could analyze all problems costlessly, and bore no costs in drafting or enforcing contracts, then all contracts would be complete.  But economists have also considered settings in which completeness is feasible but not optimal. These are cases in which information at the time of contracting is incomplete and new information is anticipated in the future.

  \begin{itemize}
    \item \emph{planned renegotiation}: rather than writing a complete contract based on incomplete information at the start of a contracting relationship, contract designers choose to write an incomplete contract, which they expect to renegotiate in the future once more information becomes available \citep{Bolton:2010,Halonen:2013}  
    \item \emph{optimal completion of contract by third-party}:rather than writing a complete contract based on incomplete information at the start of a contracting relationship, contract designers choose to write an incomplete contract, which they expect to have filled in by a third-party adjudicator with better information in the future\citep{Hadfield:1994,Shavell:2006}
\end{itemize}

The case in which a designer has to choose between developing a more complete reward structure today and deferring decisions about how to build rewards until more information has been learned seems to us to be a natural one for AI.  The question raised is about the size and nature of the risks of misalignment of releasing of an AI system into the wild when delay could allow the acquisition of better information for reward design. The problem labeled ``safe exploration" by \cite{Amodei:2016} seems to fit this description.

Finally, economists, and many legal scholars, have also proposed that contracts may be incomplete because of strategic behavior
\begin{itemize}
    \item \emph{strategic protection of private information}: a party with private information about a missing contingency does not prompt contracting to cover the contingency because doing so will reveal private information that reduces the value of the contract \citep{Spier:1992, Ayres:1989}
    \item \emph{deterring strategic investments in costly cognition} : the parties choose not to cover all contingencies because learning about them would be biased and wasteful, partly motivated by efforts to protect against strategic wealth transfers that will occur if the contingency arises \citep{tirole2009cognition}
    \item \emph{strategic ambiguity}:the parties choose not to include all known and contractible contingencies in order to control strategic behavior in response to other noncontractible contingencies \citep{Bernheim:1998}
\end{itemize}
We don't think of machines and human designers bargaining over the machines's reward function as humans bargain over contract terms and so these accounts of strategic interaction do not translate as obviously to the AI context. But strategic considerations on the part of human designers could still lead designers to choose to develop agents that are deliberately not given a complete specification of everything the designer cares about. This type of technique is common in the domain adaptation literature, which tackles the problem of what to do when you have a small amount of data from the setting/distribution that is of interest (such as the behavior of an object in the real world) but can obtain a lot of data for training purposes from a different setting/distribution (such as the behavior of a computer-generated object in a simulated environment). \cite{Ajakan:2015}, for example, present a representation learning algorithm that achieves better results than standard neural networks in generalizing classifications into a new testing domain if information about the training domain is hidden from the network; that is, the learning algorithm is intentionally incomplete.  One could think of this as strategic behavior on the part of the designer, intended to overcome distortions in the learning algorithm that will otherwise arise. \cite{Amodei:2016} call this ``adversarial blinding". Strategic incompleteness in reward design may also become relevant in more advanced systems than those we have today \citep{armstrong2015motivated} if we contemplate sequential reward design with a powerful agent.  If a robot predicts that the human may rewrite the reward structure, for example, then the robot, currently implementing the initial reward, may behave strategically--withholding information--to influence the rewriting so as to preserve the initial reward structure. 

 Figure \ref{reasons}summarizes the parallels between the reasons for contractual incompleteness and the reasons for reward misspecification.

 
 \begin{table}[t]
     \centering
     \begin{tabular}{|l|l|}
     \hline
     \textbf{Why are contracts incomplete?} & \textbf{Why are rewards misspecified?}\\
     \hline
     Bounded rationality &  Bounded rationality \\ 
     (can't think of all contingencies) &  (negative side effects) \\ \hline
    Costly cognition/drafting & Costly engineering/design \\ \hline
    Non-contractibility & Non-implementability \\ 
    (variables not describable/verifiable to enforcer) &  (unsolved learning problems) \\ \hline
    Planned renegotiation & Planned iteration on rewards \\ \hline
    Planned completion by third party in event of dispute & Planned completion by third party \\ \hline
    Strategic behavior  & Adversarial blinding, reward preservation\\ \hline
     \end{tabular}
     \vspace{10pt}
     \caption{Parallel reasons for incompleteness and misspecification}
     \label{reasons}
 \end{table}


Our principal observation is this:  reward misspecification is not an accident; it is routine, predictable, and largely unavoidable. Optimal reward design \citep{Barto:2009, Singh:2010, Hadfield-Menell:2017b} is thus a central task for AI alignment in the same sense that optimal contract design is a central task for economics. 

\section{Insights for AI Alignment from the Economics of Incomplete Contracting}

In this section, we provide a brief overview of key results in the economic theory of incomplete contracting to identify potential insights for AI researchers.   

 At the outset, we note a seemingly critical difference between the AI alignment problem and the incomplete contracting problem.  Human agents come with their own utility functions.  They are therefore modeled by economists as strategic actors, who seek to promote their own utility and need to be incentivized to put the principal's interests ahead of their own. In contrast the context in which AI agents are designed is ultimately cooperative \citep{Dragan:2017,hadfield2016cooperative}:  An AI agent and its human designer share, in principle, the same goal.  That goal is set by the human designer. An AI agent, it would seem, is not a strategic actor that needs to be incentivized to behave well.
 
But as \cite{Omohundro:2008} first observed, an AI's goal is established by its \emph{stated} reward function and this may not be the same as the human's true goal. Any divergence between robot rewards and human values can in theory be overridden by human developers. But once an AI has been initialized with a particular reward function and training (and perhaps deployment) begun, the potential for a conflict of interest and thus strategic behavior arises. 

The potential for AIs to strategize in order to achieve goals as embodied in their initial design has been a focus of the study of superintelligence \citep{Bostrom:2014, Omohundro:2008,soares2015corrigibility}, envisioning the potential for what we will call \emph{strongly strategic} AI systems to rewrite their reward functions, alter their hardware, or manipulate humans. But the value of a strategic formulation of AI behavior does not arise only in these futuristic settings.  It arises in any routine setting in which there is a divergence between the AI's stated reward function and true or intended human value \citep{Hadfield-Menell:2017a,Hadfield-Menell:2017b}. With this divergence in rewards, the AI and the human are inherently engaged in the strategic game of each trying to take actions that allow them to do well against their own reward function, even if that reduces value for the other. We will call these systems \emph{weakly strategic}. 

\subsection{Weakly Strategic AI}
We begin by highlighting two key results from the economic literature that we think can contribute to problems in existing, weakly strategic, AI systems.
\subsubsection{Property Rights}
A core result in the early incomplete contracting literature is that when complete contingent contracts are not possible, the joint value produced by two economic actors may be maximized through the allocation of property rights over productive assets \citep{HartMoore:1988,Grossman:1986}. In particular, joint profit is maximized when property rights are allocated to the actor whose non-contractible \textit{ex ante} decisions have a bigger impact on joint profit. This is because someone has to be given authority to decide what happens when a contract does not. Optimal allocation of property rights weighs the cost of strategic behavior under alternative allocations of this authority. If the agent's non-contractible decisions have a bigger impact on the joint profit generated with an asset (such as a firm) than the principal's, for example, then the best solution might be the simple one: sell the firm to the agent. \

From an economic point of view, the allocation of property rights is just a means of determining a reward function.  In particular, an owner's reward function consists of the profit that the owner can generate by making decisions about how to use an asset. Granting "property rights" to an AI, then, could be understood as imbuing an AI with the reward function associated with the ultimate rewards generated by a productive activity in which it participates. 

Seen in this light, the insight from the analysis of property rights and incomplete contracts is one that is already at the heart of the AI alignment challenge.  "Selling the firm to the agent" is the equivalent of endowing a robot with the true human utility function.   Instead of working only to refine a reward function for a specific task, designers should perhaps also be exploring ways to incorporate information from the global returns to which that task contributes. If an AI system is being developed to tag photographs on a social media network, for example, and it is inevitable that the reward design will be incomplete, should the AI also be rewarded for the size of the network, ad revenues, success in recruiting or retaining employees, and/or negative publicity in target media? Google's recent experience with  algorithms tagging black people as gorillas \citep{Ananny:2016} probably had negative consequences on the margin for all of these metrics--because it offended widely shared values. Google as a corporation (also an artificial agent) cares about those values through the mechanisms not only of the ethics of its employees and shareholders, but also through the operational impact that transgression has on consumers, investors, advertisers, and employees. 

Indeed, in many settings, we are seeking to develop artificial agents that do well against a complex combination of objectives, not all of which can be specified. Many of these could productively be viewed through the lens of what it would mean to ``sell the firm" to the agent. The challenge of building fair algorithms, for example, is a challenge of how to endow an artificial agent with more complete utility information reflecting the fact that humans evaluate decisions like deciding who gets a job in a complex way. Fair algorithms may need to be built with a view to incorporating much more information about human valuation. Similarly, research suggests that optimizing short-term engagement, measured by click rates, may damage long-term engagement with a social network--for many reasons--as Facebook recently acknowledged \citep{Ginsberg:2016}. "Selling" Facebook to its news feed algorithms would mean figuring out how to endow those algorithms with the broad set of values that Facebook users, advertisers, and others--and hence Facebook--care about.

\subsubsection{Measurement and Multi-tasking}
An important reason that contracts are incomplete is the difficulty of specifying how an action is to be measured or conditioning payoffs on a particular measurement.  A key problem arises when an agent engages in multiple tasks, and the tasks are differentially measurable or contractible.

\cite{holmstrom1991multitask} and \cite{Baker:1994} show that the optimal incentive contract for a task that can be measured should take into account the impact of those rewards on effort put towards tasks that cannot be measured. In particular, it may be better to reduce the quality of incentives on the measurable task below what is feasible, in order to reduce the distortion introduced in the unmeasurable task. \cite{holmstrom1991multitask} give the example of paying a teacher a fixed salary rather than one contingent on students' (easily measurable) standardized test scores in order not to cause teachers to "teach to the test", spending more time on test prep and less on harder to measure teaching goals such as creative problem-solving. \cite{Baker:1994} give the example of an auto repair shop rewarding mechanics for completed repairs and thereby inducing mechanics to mislead customers about the need for repairs:  completion of repairs is easily measurable; the reliability of a mechanic's diagnosis of car problems is not. More generally, sometimes it is better for a contract not to include easily contractible actions in order not to further distort incentives with respect to non-contractible actions.\

The problem of incompleteness in reward specification for an AI system can be modeled as a problem of multi-tasking. Alignment problems routinely arise because a designer conceives of a task--get coffee--as a single task when it is in fact multiple--get coffee and avoid harming humans in the break room. Moreover, even when it is well understood that a task is complex--like driving a car--the multiple tasks are differentially capable of being incorporated into a reward function.  Arriving at a destination as fast as possible without speeding or crashing may be reasonably rewarded; accommodating other drivers to facilitate traffic flow is difficult to reward.  

The lesson of \cite{holmstrom1991multitask} for AI is that a singular focus on improving performance on the measurable task may degrade performance on the unmeasurable.  Sub-optimal rewards for speed in a self-driving vehicle, for example, may be necessary if much more unreliable rewards for cooperative driving are to have maximum effect.  And in some cases, it may be appropriate \emph{not} to include rewards for what seems to be an easily measurable outcome if other important outcomes cannot be rewarded. Another way of framing this is to emphasize that task components may not be modular, capable of optimization separately or sequentially.
\subsection{Strongly Strategic AI}
The implications of the economic analysis of incomplete contracting for strongly strategic AI are more speculative, because we don't know how (or if) such systems will evolve. But we set out briefly lines of research that may be of interest to researchers thinking through the challenge of managing powerful AI that can act in an overtly strategic way to resist changes to its reward function, for example, or evade shut-down.  
\subsubsection{Control Rights}
Property rights over assets generalize to decision or control rights: any authority to make decisions about actions in circumstances not controlled by an enforceable contract. Allocation of decision rights has implications for joint profit-maximization because control rights affect incentives to take non-contracted actions that affect joint value. \citep{Fama:1983,Aghion:1997,Baker:1999}.

 \cite{Aghion:1997} present a model of incomplete contracting that could deepen AI analysis of the impact of human control on AI learning and behavior by modeling both robot and human as strategic actors.  Their model distinguishes between formal authority--the \emph{right} to implement one's preferred choice--and real authority--the \emph{capacity} to implement one's preferred choice in practice. They show that because of the risk of intervention, an agent has reduced incentives to acquire information about the environment if the principal has formal authority \textit{and the information needed to exercise it}. Paradoxically, the principal might be able to improve the incentives of the agent to acquire knowledge by remaining uninformed and thereby making a credible commitment not to intervene. 

Models analyzing the impact of human control on robot behavior generally base the AI's prediction of the probability of intervention on observed human behavior. (See, e.g., \cite{Orseau:2016}). The \cite{Aghion:1997} model suggests that an agent's beliefs about what the human knows also matter. This consideration presumably has implications for a strongly strategic AI's incentives to share information. One interpretation of \cite{Hadfield-Menell:2017a} is as a model not only of the incentive of a robot to disable its off switch but also of its incentive to share with a human the information needed to exercise off-switch authority. Another implication is the need for system design to take into account the tradeoffs between generating information necessary for meaningful human oversight and minimizing the impact of human information on the robot's performance. This suggests also refinements of the line of analysis in \cite{Orseau:2016}: is it possible for humans to have information about robot performance that robots ignore?

\subsubsection{Costly Signaling}
The economic literature on costly signaling looks at the ways in which contracts can be designed so as to elicit private information from agents. The work originates with a seminal paper by \cite{Spence:1973} which considered the problem facing an employer seeking to determine the ability of job market applicants. Spence shows that if an employer offers a high wage contract to those who meet or exceed a specified educational level and a lower wage contract to those who do not, prospective job applicants can be induced to sort themselves such that high ability workers accept the high wage contract and low ability workers accept the low wage contract. This result holds provided that education is sufficiently more costly for low ability than high ability applicants that they do not have an incentive to pretend to be high ability.\footnote{In this model, education can be obtained by either high or low ability applicants and does not have any impact on ability level.}

This line of analysis may have an interesting application to AI alignment. A key information problem for a human designer is to know a robot's current estimate of the reward function. This estimate can be more or less aligned with the human's true utility. The key observation is that intervention substituting the human's preferred action for the robot's will be more costly for the less aligned robot than the more aligned one. This seems to suggest that it might be possible to design systems in which the willingness of a strongly strategic AI to seek human input serves as a signal of the AI's alignment.  Sheer frequency of instances in which human input is sought may not be sufficient as a signal--this is likely exploitable by a system that determines that it can seek input on actions on which alignment is good or the marginal loss associated with shifting to the human preference is low while continuing to execute without intervention on actions with high misalignment. But more complex designs, responsive not merely to frequency but also to the substance of the actions on which input is ought, might be possible.  
 
\subsubsection{Renegotiation}
There is a basic trade-off, reproduced in the AI setting, between specifying behaviors for an agent \textit{ex ante} with incomplete information and specifying optimal behaviors \textit{ex post} once more information about the state of the world is available. This creates a risk of hold-up. Even with a contract, if it is ultimately discovered that the action called for is not optimal, there is an incentive to renegotiate. In the standard incomplete contracting setting, this means having to pay the agent to agree to shift from the old contract to a new one.  A key insight is that the provisions of the initial contract set the terms on which the new contract is bargained \citep{Hart:1988}.  

Strongly strategic AI systems may need to be effectively "bought out", incentivized to shift from a reward function they were originally given to one that a human later discovers is closer to the truth. \cite{soares2015corrigibility} and \cite{armstrong2015motivated} consider this problem, but are not optimistic about the capacity to design powerful AI systems, capable of disabling shutdown or preventing change to their rewards, that will neither resist nor manipulate changes to their rewards. Uncertainty in reward design may help here \citep{Hadfield-Menell:2017a}: an agent that is designed to estimate the reward function based on information gleaned from the environment may value human input to refine that estimate.

These results are primarily theoretical in nature, showing conditions under which an agent will be indifferent to a change in its utility function or value human input. The lesson from incomplete contracting focuses on the design of initial and subsequent rewards in a practical sense. The original reward design affects the buyout structure, and creates strategic incentives to take actions to increase the capacity to extract higher payoffs for later change. Anticipating the structure of buyout challenges can then inform the design of initial rewards to facilitate buyout. 

\section{Insights for AI Alignment from the Law of Incomplete Contracting}
The economics literature on incomplete contracts provides a systematic framework for thinking through the kinds of challenges we anticipate in the design of rewards that achieve human objectives and some suggestive avenues for reward design, as we have explored above. But the deeper insights to be gained from the incomplete contracting framework, we believe, are insights that economists have also been prompted by legal scholars and behavioral social scientists to absorb. These insights have to do with the tremendous amount of social infrastructure that is deployed to make contracts effective in practice. 

Contracts do not exist in a vacuum; they come heavily embedded in social and institutions structures \citep{Granovetter:1985}. At a minimum, they depend on shared language and organized structures for enforcement: formal enforcement through courts and coercive authorities and informal enforcement through social sanctions such as collective (coordinated) criticism and exclusion from valuable relationships. Incomplete contracts depend even more extensively on these external, third-party institutions: not only to enforce contractual terms but to supply contractual terms by interpreting ambiguous terms and filling in gaps.  They contain not only their express terms but also their \emph{interpreted} and \emph{implied terms} \citep{Farnsworth:1968}. This important point has been emphasized by legal scholars, in a field known as relational contracting, for several decades \citep{Macaulay:1963,Macneil:1974,Macneil:1978,Macneil:1983,Goetz:1981}. This concept overlaps with the economist's definition of relational contracting (see \cite{Gil:2017} for a review) to the extent that it focuses on informal sanctions for breach of obligations; legal scholars in the sociological tradition have always included here a wide variety of sanctions (including internalized sanctions such as guilt and shame) whereas economists have modeled specifically the termination of valuable economic relationships or the degradation of reputation, which reduces contracting opportunities with third-parties in the future. But the legal concept of relational contracts goes further than enforcement to focus also on the importation of obligations into a contractual relationships from sources other than the express language of an agreement. \cite{Williamson:1975}, \cite{Alchian:1972} and \cite{Klein:1978} were the first economic treatments to focus on this aspect of the legal analysis of relational contracts and \cite{Hadfield:1990} was an early effort to integrate economic concepts of relational contracting with the legal framework of implied terms.  

The lesson that \cite{Granovetter:1985} pressed on economists in the early stages of the analysis of incomplete contracting, however, seems equally apt for AI researchers tackling the problem of alignment today: alignment problems cannot be solved without support from external normative structure.

\begin {figure}
\includegraphics[scale=.3]{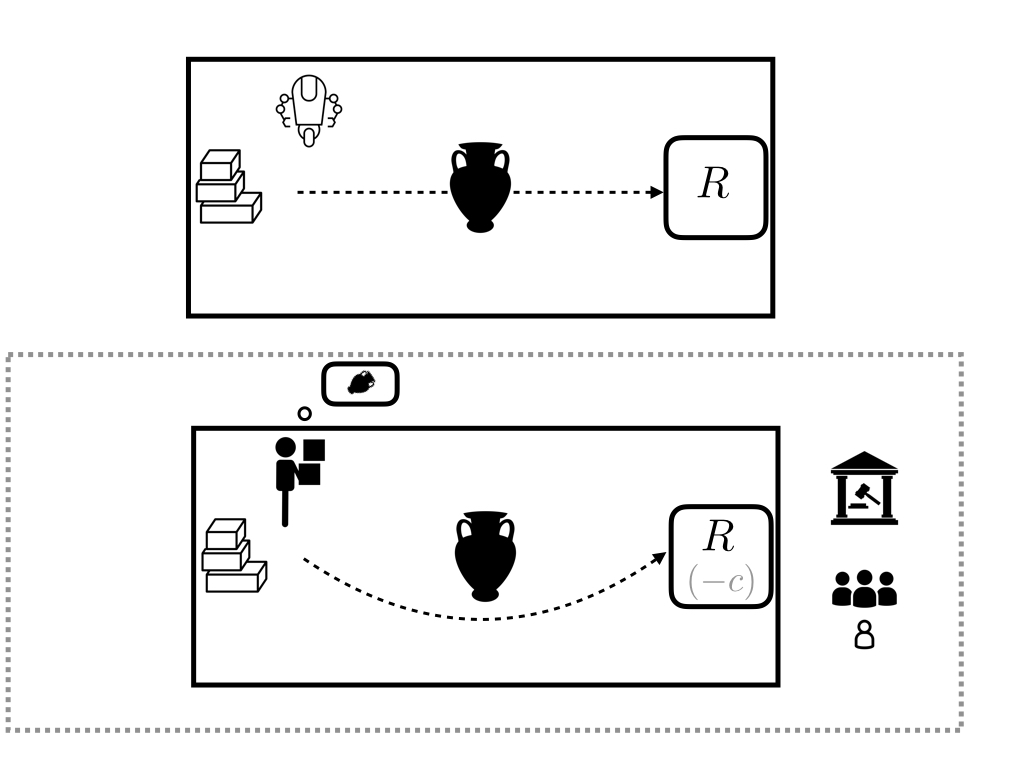}
\centering
\caption{When a robot is given a reward function that specifies a reward only based on moving boxes, it will ignore a vase that appears in the path \citep{Amodei:2016}. If a human agent is given a contract that pays only for moving boxes, she will interpret her contract to include an implied term that penalizes knocking over the vase. This is because the human contract is embedded in a normative environment in which institutions such as courts and culture impose formal sanctions (such as money damages) or informal sanctions (such as exclusion from valuable economic and social relationships) for violating formal rules and informal norms. The human agent will naturally refer to this environment to determine the true contract. Achieving a similar result for the robot will require building the technical tools that allow the robot to make a similar appeal to external sources of norms and values to address misspecification.}
\label{vases}
\end{figure}

Consider a simple example posed by \cite{Amodei:2016}: a robot learns to move boxes from one side of a room to another in a training environment that lacks obstacles. (See Figure \ref{vases}.) When deployed, a vase of water appears in the path the robot has learned to use.  The robot that is rewarded for transferring boxes and not penalized for knocking over the vase will ignore the vase. \cite{Amodei:2016} use this as an example of negative side-effects:  unintended consequences arising from a failure to include relevant features in the robot's reward.  They make the same observation that we do: it is not possible to identify all the features that might arise as the robot carries out its task. Reward misspecification is unavoidable.  They explore potential solutions such as penalizing the robot for having an impact on the environment or getting itself into situations where it can have influence over an environment. As they recognize, the problem is hard--we want the robot to be able to make changes to the environment like moving boxes--and may be very difficult to generalize across different settings.    

Consider now what happens if we hire a human agent to carry boxes. Suppose the contract is a direct analogue of the robot's reward function.  It says that the agent will be paid a certain amount for every box carried to the other side of the room. It says nothing about knocking over vases, and there are no vases about when the agent is hired. What happens when a vase appears? Easy: the agent will walk around the vase. Why?


The reason is that the incomplete agreement that says nothing about vases is not the entire contract between principal and agent. Human contracts are not limited to the express terms; they also include implied terms. In particular, in these circumstances, the contract implicitly contains a term that imposes a cost on breaking the vase.  If the agent ignores the vase, the reward the agent receives will be reduced by some amount: the agent might be charged for the breakage, she may suffer future income losses as a result of being fired or earning a bad reputation in the labor market, she may suffer psychic pain as a result of being criticized, or she may suffer the discomfort of feeling guilty, ashamed, or incompetent for having done something wrong.

These implied terms represent a type of ``common sense'' reasoning --- reasoning about the extensive normative structure in which human contracting is embedded.  Our human contract does not arise between two people who have always lived alone on a desert island. It arises in an environment filled with rules and institutions that resolve, precisely, questions such as: was it wrong for the agent to knock over the vase while carrying out this task? Some of these rules might come in the form of cultural norms--classifications that arise as emergent features from repeated interaction and discussion among participants in a group \citep{Bicchieri:2006, Wiessner:2005, Hadfield:2017}. Others of these rules are the product of formal dispute-resolution systems of law and adjudication that humans administer to fill in the gaps in incomplete contracts. When secured by an effective enforcement scheme--either centralized imposition of penalties or decentralized coordination social punishment--these rules constitute a \emph{normative social order} \citep{Hadfield:2014}. 

This normative social order completes the contract; it is a resource available to contracting members of a community that has successfully coordinated and incentivized a third-party punishment regime. Humans engaged in contracting do so with common knowledge that their contracting efforts are embedded in this social order. This makes them more willing to contract and take on the risk associated with the inevitable incompleteness of their agreement.  They both know that in the event of a gap or ambiguity, the terms of the contract will be filled in by a process that they can expect to be neutral and which they can reason through. The principal, knowing he will inevitably forget to include all the terms governing the agent's execution of the task of carrying boxes, can anticipate at the time of contracting that if things like vases later appear, the agent will be able to figure out that she should avoid knocking them over and that she will pay some penalty if she does not. Similarly, the agent can anticipate at the time of contracting that reasonable obligations, like avoiding breakage, will be treated as an implicit part of the contract but that unreasonable obligations, like finishing the job even if the principal locks her out, will not. She can consult external normative infrastructure to determine what it says about vases when they appear. This external normative infrastructure makes rational incomplete contracting possible.

We conjecture that any robust solution to the AI alignment problem will also require the recruitment of normative resources external to the reward structure designed for any particular application. 

We do not mean by this embedding into the AI the particular \emph{norms and values} of a human community. We think this is as impossible a task as writing a complete contract.  Human norms and values are highly variable and deeply contextual. \cite{Amodei:2016} anticipate this when they consider the shortcomings of generic approaches that seek to create incentives to ``minimize impact on the environment.'' However, it is easy to come up with examples of cases in which ``minimize impact'' is an inadequate formalization of the right action.  Suppose, for example, that the contract puts a time limit on the movement of the boxes and the agent can't move them all in time without knocking over the vase. Or the boxes contain medical equipment that is urgently needed to treat trauma patients and nobody cares about broken vases.  Or the vase is a sacred object and the people who need the medical treatment also believe that knocking over the vase will anger the gods and only make recovery less likely. Human societies--throughout history, across today's diverse planet, and in a highly transformed future--contain a multitude of settings in which there is no easy answer to the question of whether it is okay or not okay to break the vase.

We usually refer to adapting to these different contexts as ``common sense,'' but it is important to emphasize that this is common sense \emph{about what actions society will sanction.} The human agent's capacity to infer implied terms about the values associated with breaking the vase is a product of the human's ability to interact with and participate in normative social structure.  Humans are endowed with the cognitive architecture needed to read and predict the responses of this normative structure. The human agent avoids the vase in those circumstances in which she concludes that the community will sanction breaking it and plows on through otherwise. In some cases the conclusion will be hard to reach on her own and for that she--and the principal who hired her--may have recourse to classification institutions (like courts and lawyers) to resolve the ambiguity. 

Aligning AI with human values, then, will require figuring out how to build the technical tools that will allow a robot to replicate the human agent's ability to read and predict the responses of human normative structure, whatever its content.  To figure out, as the human does, whether the contract (reward function) also includes terms according to which the community judges particular actions not explicitly addressed by the contract (reward function) to be wrongful. Learning how to read and participate in normative social structures is a critical part of human intelligence and for AIs to be aligned with humans will require their learning how to do this also. (We leave for another day the intriguing question of whether we will also need to build specialized legal institutions and processes adapted to performing the function of resolving ambiguity for robots.)

Building AI that can supplement their designed rewards with implied rewards from community normative structure also will require building tools that allow a robot to assign negative weight to actions classified as sanctionable by the community. Human agents assign costs to taking actions that violate implied terms in contracts for many reasons.  Some of these penalties are formally imposed by enforcement institutions such as courts (contract damages, for example). Others are imposed informally through coordinated community action:  refusing to hire or do business or engage socially with someone who is judged by formal or informal standards to have breached a contract, for example \citep{Bernstein:1992, Bernstein:2015, Hadfield:2016}. These externally-imposed penalties seem difficult to transpose to the AI context.  

But humans also internalize social penalties. An important part of human development is the building of the cognitive architecture for experiencing negative emotions such as distress, shame, and guilt in response to a real or imagined public judgment of rule violation. Adam \cite{Smith:1759} famously referred to this as the capability to view one's own conduct as if through the eyes of an ``impartial spectator." Recent work in developmental psychology has shown, for example, that even very young children will spontaneously criticize the violation of arbitrary rules (of a novel game to which they have just been introduced, for example) and somewhere between two and three years old, children understand that rules can be context-specific \citep{Rakoczy:2009,Rakoczy:2013}. This cognitive architecture--creating the mental buttons that external normative criticism can press to change behavior, so to speak--would seem to have a natural analog in the design of an artificial intelligence--assigning loss to conditions in which the prediction is made that an action, in context, would be judged by external human communities to be wrongful.

Building AI that can reliably learn, predict, and respond to a human community's normative structure is a distinct research program to building AI that can learn human preferences.  Preferences are a formalization of a human's subjective evaluation of two alternative actions or objects. The unit of analysis is the individual. The unit of analysis for predicting the content of a community's normative structure is an aggregate: the equilibrium or otherwise durable patterns of behavior in a group. A solidly microfoundational account of community normative structure will build on individual perceptions, valuations and actions \citep{Hadfield:2014}. But the object of interest is what emerges from the interaction of multiple individuals, and will not be reducible to preferences.  Indeed, to the extent that preferences merely capture the valuation an agent places on different courses of action with normative salience to a group, preferences are the outcome of the process of evaluating likely community responses and choosing actions on that basis, not a primitive of choice.

\section{Conclusion}

The alignment of artificially  intelligent agents with human goals and values is a fundamental challenge in AI research. It is also the fundamental challenge of organizing human economic interaction in an economy built on specialization and the division of labor--in which humans are tasked with taking actions that generate costs and benefits for other humans. By recognizing and elaborating the parallels between the challenge of incomplete contracting in the human principal-agent setting and the challenge of misspecification in robot reward functions, this paper provides AI researchers with a different framework for the alignment problem. That framework urges researchers to see reward misspecification as fundamental and not merely the result of poor engineering. Doing so, as we show, generates insights both for the analysis of current, weakly strategic, AI systems and potential, strongly strategic, systems. Our most important claim is that aligning robots with humans will inevitably require building the technical tools to allow AI to do what human agents do naturally: import into their assessment of rewards the costs associated with taking actions tagged as wrongful by human communities. These are the lessons learned by economists and legal scholars over the past several decades in the context of incomplete contracting. They are lessons available also to AI researchers.


\bibliographystyle{icml2017}
\bibliography{incomplete_contracts}


\end{document}